\documentclass[runningheads]{llncs}
\usepackage[T1]{fontenc}
\usepackage{graphicx}
\usepackage[ruled,vlined]{algorithm2e}
\usepackage{hyperref}
\usepackage{booktabs}
\usepackage{float}
\usepackage{amsmath}
\usepackage{caption}
\usepackage{subcaption}
\usepackage{tikz}
\usetikzlibrary{shapes,arrows,automata, positioning}

\usepackage{algorithmic}
\usepackage{tabularx}
\usepackage{bm}
\usepackage{amssymb}
\usepackage{booktabs}
\usepackage{multirow}

\usepackage[textsize=tiny, backgroundcolor=orange!20!white]{todonotes}
\usepackage{xcolor}
\usepackage{soul}

\begin{document}
\title{DriftGAN: Using historical data for Unsupervised Recurring Drift Detection} 
%
%
\author{Christofer Fellicious\inst{1}\orcidID{0000-0001-7487-7110} \and 
\\Sahib Julka\inst{1}\orcidID{0000-0002-2952-5519} \and \\Lorenz Wendlinger\inst{1}\orcidID{0000-0001-9459-6244} \and \\
Michael Granitzer\inst{1}\orcidID{0000-0003-3566-5507}}

\authorrunning{Fellicious et al.}
%
\institute{University of Passau, Germany\\
\email{fistname.lastname@uni-passau.de}\\}
\maketitle              
\begin{abstract}
In real-world applications, input data distributions are rarely static over a period of time, a phenomenon known as concept drift. Such concept drifts degrade the model's prediction performance, and therefore we require methods to overcome these issues. The initial step is to identify concept drifts and have a training method in place to recover the model's performance. 
Most concept drift detection methods work on detecting concept drifts and signalling the requirement to retrain the model. However, in real-world cases, there could be concept drifts that recur over a period of time. In this paper, we present an unsupervised method based on Generative Adversarial Networks(GAN) to detect concept drifts and identify whether a specific concept drift occurred in the past. Our method reduces the time and data the model requires to get up to speed for recurring drifts. Our key results indicate that our proposed model can outperform the current state-of-the-art models in most datasets. We also test our method on a real-world use case from astrophysics, where we detect the bow shock and magnetopause crossings with better results than the existing methods in the domain.

\end{abstract}
\section{Introduction}\label{sec:introduction}
Machine Learning is ubiquitous in our lives and has the potential to revolutionize many industries by increasing efficiency, personalizing experiences, predicting outcomes, automating tasks, and improving decision-making. Machine learning often predicts different aspects of our lives, such as the prediction of energy consumption~\cite{seyedzadeh2018machine} or the fuel consumption of trucks~\cite{perrotta2017application}. For applications with continuous stream of data that span months or years, there is a good chance for the input data distribution to change over time. For example, in credit card fraud detection, customers' buying behavior changes over time~\cite{ghosh2022nag}. Such changes in the input data distribution are known as \emph{concept drifts} and adversely affect the model's prediction performance.
To counter such performance degradation, we need mechanisms to identify the changes in the input data distribution and retrain the prediction model so that the prediction model adapts the changed input data distribution and improves the performance.
In the most straightforward strategy, we assume changes happen regularly and train the model periodically. However, this naive strategy comes with a couple of significant disadvantages. First, model retraining is usually very costly in terms of time for data collection and computation, and it is even harder to determine the perfect retraining interval. Secondly, changes in distribution may not coincide with the periodic model retraining schedule, thereby rendering model update efforts inefficient and ineffective. Therefore, we need a mechanism to detect the changes in the underlying data distribution and retrain the model for optimal performance and utilization of computing resources~\cite{gama2004learning}.

However, detecting changes in data distributions and identifying previous data distributions hold the potential to improve classifiers, which is still relatively unexplored. 
In particular, there could be patterns that reappear later. For example, suppose we have a method to identify such past data distributions. In that case, identifying recurring changes in the distribution will help us bolster the training set and get the model up to speed quicker rather than waiting for the required number 
of instances from the real-world data stream. Currently, most concept drift algorithms only focus on identifying the changes in the distribution, but do not consider whether a similar data distribution previously occurred. Examples of repeating distributions are seasonal repeating customer shopping behavior, energy consumption, or changes in credit card fraud patterns. 

In standard methods, we need to wait until enough input is acquired to train the model after every new concept drift. Furthermore, this data also needs to be labelled, which increases the cost and energy expenditure.
Contrary to standard methods, our framework keeps track of the previously seen data distributions, which allows us to augment the retraining dataset with the previously seen data of the same distribution. In real-world cases, this augmentation helps reduce the time required to acquire enough new data on the new distribution, thereby reducing the model's turnaround time. As a result, we also increase the performance of the model when compared to retraining the model from scratch with limited data.
This time to obtain the labels and retrain the model is significant when labeled data is either sparse or only available with a significant time delay~\cite{dastidar2022importance}. 

In this work, we extend drift detection by keeping track of all data distributions seen by the model. 
We show that using the data from the past concept drifts for retraining the model allows us to outperform the current state of the art. Furthermore, this
memory of previous data distributions allows the model to adapt and reduce the turnaround time to previously identified concept drifts. Methodologically, we develop a multiclass-discriminator drift Generative Adversarial Network (DriftGAN). The DriftGAN contains a discriminator module that does not only distinguish between real and artificial examples but also discriminates to which previous drift the current example belongs. Since drifts can increase over time, the discriminator network can grow incrementally by adding more discriminator classes whenever we identify new data distributions. We evaluate our DriftGAN on eight unsupervised real-world drift detection datasets and compare it with existing unsupervised methods. 
Unsupervised drift detection does not rely on the performance  of the label prediction model to detect drift - which is not readily available in most tasks 
but only based on changes in the input distribution. The results show that our method has a better performance when compared to other unsupervised methods that only detect whether drifts occur or not. We also adapted our method as a sampling strategy for the magnetometer data of the NASA MESSENGER spacecraft. Our results for the real-world application scenario also produce better results than the existing methods, which shows that our method can be adapted to different domains with very little to no changes.


We structure the rest of the paper as follows,~\autoref{sec:related_work} describes the related work with supervised and unsupervised concept drift detection,~\autoref{sec:method} discusses our approach to the problem with the necessary explanation on the identification of previously seen data distributions and the research gap we address. 
Then,~\autoref{sec:results} describes our experimental results and the interpretation of the results. Finally,~\autoref{sec:conclusion} concludes our paper.
\section{Related Work}\label{sec:related_work}
We can broadly classify the concept drift detection methods into Supervised and Unsupervised methods. 
Supervised methods have access to the input feature vector's label, making it easier to understand and detect data distribution changes. The problem becomes compounded with unsupervised learning methods as the data labels are unavailable to the model to decide whether the distribution has changed. Unsupervised learning methods are more useful in real-world scenarios as data labels might not be readily available to the model and may require time or expertise to obtain. 

\subsection{Supervised Methods} 
The Drift Detection Method (DDM) is one of the earliest supervised methods in concept drift detection~\cite{gama2004learning}. The method uses the error rate to determine whether a concept drift occurred. The idea behind this method is that the error rate increases when there is a change in input data distribution. The algorithm defines two thresholds based on the error rate. The first threshold is also lower, serving as a drift warning. The second threshold signals the drift. The retraining occurs in the window that starts from the drift warning.

The Adaptive Windowing (ADWIN) method uses different-sized windows to detect concept drifts~\cite{bifet2007learning}. ADWIN uses statistical tests on different windows and detects concept drifts based on the error rates, similar to DDM. Cabral and Barros proposed a method using Fisher's Exact Test~\cite{de2018concept}. This method analyzed two separate context windows and the assumption that if there was no drift, the error distribution should be equal between the two windows. They use the generalized contingency table from Fisher's Exact test to test the null hypothesis.

\subsection{Unsupervised Methods}

Identifying concept drifts without the corresponding labels to the input feature vectors is comparatively more difficult. We can only rely on the input feature vectors to decide whether a concept drift occurs. Additionally, unsupervised concept drift detection cannot detect certain types of concept drifts such as changes in the drift label while the input data distribution is more or less constant. In such scenarios, it is almost impossible to identify a concept drift without the class labels. Hu et al. explain why detecting concept drifts while maintaining high performance is not trivial~\cite{hu2020no}. The paper explains high performance with labeled data is possible, but labeling data streams as mentioned earlier could be expensive and time-consuming. These drawbacks of supervised drift detection mean that even with all the drawbacks, unsupervised drift detection is more efficient in real-world scenarios to detect concept drifts.

The Kolmogorov-Smirnov is a nonparamtetric statistical test can determine whether two data points belong to the same distribution. The problem with the K-S test is its time complexity that grows with the number of samples. Reis et al. proposed a faster alternative to the K-S test using an incremental K-S test~\cite{dos2016fast}. This implementation uses a custom tree data structure called "\emph{treap}." The algorithm uses two windows, where the first is fixed and the second is sliding. The first window contains the original concepts the model was trained on, while the second one slides and checks whether a concept drift occurred. While the K-S test remains popular for univariate data and an expansion of the K-S test for multivariate data was proposed by Justel et al.~\cite{justel1997multivariate}.

Discriminative Drift Detector is an unsupervised method for detecting concept drifts using a linear regressor on two windows~\cite{gozuaccik2019unsupervised}. The basic idea is that there are two windows; the first is assumed to be from the old data distribution and given label \textbf{0}. A second window, a sequence of input data, is chosen and labeled as \textbf{1}. If the linear regressor can discriminate between both windows with a certain precision means that the two windows are different, and thus a concept drift has occurred. The performance is measured using a secondary supervised Hoeffding Tree classifier. 
One Class Drift Detection(OCDD) is a method that uses a linear SVM and sliding windows to detect outliers in the incoming data stream~\cite{gozuaccik2021concept}. When the outliers reach a threshold, the algorithm signals a concept drift. Currently, OCDD has the best performance in unsupervised drift detection.

Another method, ODIN, uses autoencoders to identify drifts and select trained classifiers based on the drift~\cite{suprem2020odin}. ODIN consists of four components; an encoder, a decoder, a latent discriminator and an image discriminator. Autoencoders generate latent representations from input feature vectors, and a decoder generates the initial feature vectors. The latent representation is passed to the latent discriminator for detecting inlier frames from outlier frames. The loss function is computed as the weighted sum of the loss of the latent discriminator loss, the image discriminator, and the standard reconstruction loss. 

Methods like the K-S test are unsuitable for data with many features, as checking the data distribution across each feature is not feasible. All methods in the concept drift domain are adept at identifying supervised and unsupervised drifts. However, they do not consider the possibility of recurring drifts and use the historical data from previous drifts to retrain the model. Reusing previous data from the same distributions that occurred in the past will enable us to retrain the model faster and reduce the false predictions until a new model is trained. 

Gulcan and Can proposed a method that addresses concept drift in multi label data streams~\cite{gulcan2023unsupervised}. The method Label Dependency Drift Detector(LD3) works alongside any multi-label classifier. LD3 works by exploiting the predicted label dependencies and if there is a significant change in the dependencies, a concept drift is signalled. The algorithm uses a threshold and two windows, one for new samples and one for old samples, along with a correlation window that keeps track of past correlations between the data points. 

Jain et al. address the issue of concept drift in anomaly detection of network traffic~\cite{jain2022k}. The authors propose two techniques, an Error Rate based concept drift detection and Data Distribution based Concept Drift Detection. The method uses a K-Means Clustering for reducing the dataset size and upgrade the training dataset. The final classifier is a Support Vector Machine(SVM) for anomaly detection.
\section{Method}\label{sec:method}
Generative Adversarial Networks(GAN) are valuable for creating synthetic data. Goodfellow et al. introduced the concept of generative adversarial networks~\cite{goodfellow2014generative}. A GAN comprises a \textbf{Generator} and a \textbf{Discriminator}. The \textbf{Generator} creates synthetic data, and the \textbf{Discriminator} predicts whether the input data belongs to either the real or synthetic data created by the Generator. The training continues until the discriminator's loss drops below a certain threshold. Generative Adversarial Networks are used in a wide variety of domains such as for music generation~\cite{kumar2019melgan}, image super-resolution~\cite{zhang2019ranksrgan}, or even medical image reconstruction ~\cite{chen2018efficient}. We understand from the different domain applications that Generative Adversarial Networks are good at generating different output distributions. Furthermore, this is the property we use in our method.~\autoref{sec:method} describes the complete method. The code is available as open source\footnote{\url{https://github.com/cfellicious/DriftGAN}}.

\subsection{Framework}

We use a \textbf{\textit{Generator}} $\theta_{G}(s)$ that accepts a sequence $s=<x_{t-k}\ldots x_{t}>$ with length $k$ of input feature vectors $x$. The main aim of the Generator is to predict the next feature vector $x_{n+1}$. More precisely, in~\autoref{eqn:generator} we define $\theta_{G}(x)$ as 

\begin{equation}
    \begin{aligned}
        \theta_{G}(s): x_{n-k}, ... ,  x_n \longmapsto x_{n+1}
    \end{aligned}
    \label{eqn:generator}
\end{equation}

The \textbf{\textit{Discriminator}} $\theta_D(x)$ is a multi-class dynamic discriminator that predicts whether an input data belongs to one of the input distributions from the data or is a synthetic data point generated by $\theta_{G}(s)$.
We define a mapping function $\theta_{D}(x)$, which maps the incoming feature vectors to a set of identified input data distributions or an unidentified data distribution, as explained in ~\autoref{eqn:discriminator}.
\begin{equation}
\begin{aligned}
    \theta_{D}(x): x_i \longmapsto {\Pi_1, ....,\Pi_n, \Pi_{n+1}},\\
\end{aligned}
\label{eqn:discriminator}
\end{equation}
\textit{where},
\vspace{-0.2cm}
\begin{itemize}
    \item $x_i$ is an input feature vector
    \item $\Pi_j$ is the $j-th$ input distribution with $j=1$ indicating the current distribution and $j>1$ being the $j-1$ past distribution detected by our GAN
\end{itemize}


\paragraph{Detecting Drifts:}
We map all incoming input feature vectors to one of the distributions. 
We signal a drift if all the contiguous inputs in a batch are mapped to any of the same data distributions other than the current data distribution($P_1$). \autoref{eqn:drift_equation} gives the formal description.

\begin{equation}
\begin{aligned}
    \forall x \in \delta, where \theta_D(x) ~ \ne ~ \Pi_{curr} ~ ~\& \\~ ~ \forall(x_1,x_2)~\in~\delta,~\theta_D(x_1)=\theta_D(x_2)
\end{aligned}
\label{eqn:drift_equation}
\end{equation}
\textit{where},
\begin{itemize}
    \item $x$ is an input feature vector
    \item $\delta$ is the input data batch to $\theta_D$
\end{itemize}

\paragraph{Dynamic Discriminator Extension:}
Initially, we start with a training dataset of size $\rho$ and train the GAN so that $\theta_D$ learns to discriminate the real data from the data generated by $\theta_G$. The current data distribution is the data distribution of the training dataset of size $\rho$. Data generated by $\theta_G$ represents the drifted data.
$\theta_G$ can generate the next in the sequence from completely different input distributions, and can also generate data sequences that do not belong to any previously seen data distributions. This capability of $\theta_G$ allows us to set a class for previously unseen distributions in $\theta_D$. We also label the data so that anything other than previously seen distributions is assigned to the same class.
A concept drift can be either from a previously seen distribution, except for the first concept drift, or an entirely new drift. If a previously seen data distribution repeats, we use the data from that distribution to bolster the training of the secondary classifier. It is also possible to reuse only the old data to train the classifier or even the old classifier itself if it is stored. However, we do not choose to do so because there could be different data distributions in an extensive dataset, and the space requirements might be infeasible for larger datasets.

\begin{table*}
\centering
\caption{Hyperparameters used in the code, their symbols and explanations}
  \begin{tabular}{cll}
    \toprule
    Symbol & Hyperparameter & Explanation\\
    \midrule
    $\rho$ & Window Size & Number of training instances for drifts\\
    $\delta$ & Batch size & Number of testing instances that should agree for a drift  \\
    $W$ & Data Window & Window of sequential input feature vectors of size $\rho$ \\
    $\delta$ & Data Batch & A batch of data for testing \\
    $\Pi(x)$ & Input distribution & Represents the input distribution of $x$ \\
    $\Pi_{curr}$ & Current distribution & Drift not signalled as long as input in this distribution\\
    $\omega$ & Seen distributions & Set of seen input distributions  $\Pi(x_{n}) \in \omega$\\
    $\mu$ & Mean & Mean of the input feature vector \\
    $\sigma$ & Standard Deviation & Standard deviation of the input feature vector \\
    $\lambda$ & Fraction & Percentage of historical data to be used for training\\    
    \bottomrule
\end{tabular}
\label{tab:hyperparameters}
\end{table*}

\begin{algorithm*}
\caption{GAN-based Drift Detection.~\autoref{tab:hyperparameters} explains the hyperparameters}
 FindDrifts($\rho$, $b$)  $\rightarrow$  Window size and batch size respectively \;
 
 $W = data[1:\rho]$ \hspace{1cm}     $\rightarrow$ Obtain $\rho$ samples for training the GAN\;
 $\theta_{G}, \theta_{D}$ = TrainGan($W$) \hspace{1cm}     $\rightarrow$ Train the GAN\;
 $\Pi_{curr} = \Pi(W)$  $\rightarrow$ Set the current distribution\;
 $\omega = {\Pi_{curr}}$ $\rightarrow$ Add curr distribution to seen distributions
 \While{isDataBatchAvailable}{
 $W = x_{n-b} .. x_b $  $\rightarrow$ Obtain new batch \;
 $\delta = \forall x \in W (x - \mu(x))/\sigma(x)$ \;
 \If{$\forall x \in \delta, where \theta_D(x) ~ \ne ~ \Pi_{curr} ~ ~\& ~ ~ \forall(x_1,x_2)~\in~\delta~\theta_D(x_1)=\theta_D(x_2)$}{
    Batch prediction consensus on input not belonging to current data distribution, Drift Detected \;
    \If{$\Pi(\delta) \notin \omega$}{
        New input distribution detected  \;
        $\omega = \omega \cup \Pi(\delta)$ \;
        Retrain $\theta_D$ using $\theta_G$ with the added new input distribution $\Pi(\delta)$\;
        }
        $\Pi_{curr} = \Pi(\delta)$
  }
 }
\label{algo:code}
\end{algorithm*}

One of the main issues with Generative Adversarial Networks is the mode collapse. Mode collapse occurs when the Generator learns only one of the input modes instead of generating generalized data.
We counter the mode collapse problem by having a random mini-batch input to the Generator ($\theta_G$) and the Discriminator($\theta_D$). Using mini batches forces the generator network to learn to generalize better rather than falling back to a single mode in the input distribution. We also standardize the input data by the mean and standard deviation, which compresses the input feature vector. 
ODIN implemented a variation of the standardization method~\cite{suprem2020odin}. 
\subsection{Network Architectures}

For our framework, we need two networks, one for the \textbf{Generator}($\theta_G$) and one for the \textbf{Discriminator}($\theta_D$). The Generator predicts the input feature vector in the sequence, and therefore we can get away with using a smaller network to learn the function. On the other hand, the multi-class Discriminator requires much more expressive power and therefore has a slightly larger network when compared to the Generator. 
The symbols $N, P, R, S$ represent the number of neurons in each layer. In our case, $N$ equals the sequence length multiplied by the input feature vector size, and $P$ and $R$ are constants of 128 and 4096, respectively. $S$ is the output size equal to the size of a single input feature vector.

\tikzset{%
  every neuron/.style={
    circle,
    draw,
    minimum size=1cm
  },
  neuron missing/.style={
    draw=none, 
    scale=4,
    text height=0.333cm,
    execute at begin node=\color{black}$\vdots$
  },
}

The $Discriminator$($\theta_D$) is a slightly larger network, and the size of the top layer of the network is incremented by one every time a previously unseen drift is detected. The first three layers do not change during the execution. We have two layers of 1024 neurons, each with the final layer having a Sigmoid non-linearity to the final layer. The number of neurons on the final layer at any point depends on the number of unique drifts seen by the discriminator plus one neuron as a generalization of all the unseen drifts.

The main objective on the design of the GAN is to have the networks as small as possible as the GAN is not a static part of the detection process. Every time a concept drift is detected, the GAN undergoes training and having a larger model that could capture and learn more information could be detrimental as it takes longer to train. Therefore, we avoid using larger and more complex networks and go for the simplest methods to speedup the training. The primary reason that small networks work well in our experiment setting is due to the fact that we do not use noise as the latent vectors to $\theta_G$ but rather a sequence of real input. This allows some leeway with having a smaller network to SACadapt to the inputs to only predict the next in the sequence.

\subsection{Training}
We start with an initial window of size $\rho$ where the input feature vectors of that window are assumed to belong to the same distribution. The Discriminator, at this point, has only two output neurons, one for the current distribution of the window and one for unseen distributions. We use the Adadelta optimizer for both the Generator and discriminator. Furthermore, we use the CrossEntropyLoss function for the Discriminator. As for the Generator, the loss is a Mean Squared Error loss based on the actual $n+1$ vector in the sequence added to the CrossEntropyLoss of the Discriminator. These additions of loss values allow the Generator to fool the Discriminator better while also better predicting the next in-sequence input vector.
\section{Results}\label{sec:results}
We evaluate our approach on different real-world datasets and compare it with the current state of the art. We also use our approach to predict the incoming data distributions and predict on the different magnetic events of the MESSENGER spacecraft while orbiting Mercury.

The current state of the art for unsupervised drift detection is from~\cite{gozuaccik2019unsupervised} on a public dataset.
We use the same datasets used by the authors for the state-of-the-art.
In order to facilitate a direct comparison to state-of-the-art, we use the same real-world datasets, the same metrics, and the secondary classifier for our experiments. Using the same datasets allows us for an apples-to-apples comparison.

The secondary classifier predicts the actual labels on the incoming data. The better the concept drift identification of our model, the better the performance of the secondary classifier, which is a supervised learning algorithm, a Hoeffding Tree Classifier, in our case. We use a Hoeffding Tree Classifier as the secondary supervised classifier because of the support for streaming datasets. The main reason behind the choice is to enable better comparison with the state of the art as they use the same classifier as their secondary classifier.

Our comparison is based on the results of the current state-of-the-art in unsupervised concept drift detection published by~\cite{gozuaccik2019unsupervised}. We use accuracy as the metric for evaluating all the datasets as it is the most prevalent metric for analyzing the algorithm's performance. Most other publications also use the accuracy metric; hence, we use the same metric to evaluate our experiments for comparison. We use the implementation of scikit-learn~\cite{pedregosa2011scikit}. In addition, we use the implementation of the scikit-multiflow library for the Hoeffding Tree classifier~\cite{montiel2018scikit}. This library caters to analyzing streaming datasets. For the evaluation, we use the Interleaved-Test-Then-Train method outlined in the paper by~\cite{gozuaccik2021concept}. This method tests every incoming feature vector and then adapts the classifier to the tested feature vector. We reset and retrain the Hoeffding tree classifier whenever a concept drift is detected.

\subsection{Standard Drift Datasets}

We compare our results with the state of the art but we also developed three baseline methods. The baseline methods are explained below.

\subsubsection{Baseline Methods:~~}

In addition to comparing our results with state of the art, we also compared the experimental results with three additional methods that we call our baseline methods for the experiment. The baseline methods we selected are probable real-world scenarios where concept drift is not considered. The first method is one we see in most machine learning models where the model is trained on a training set and then never modified. The purpose of such a baseline is to provide an understanding of the performance if we never consider concept drift and let the model predict over the whole test set. We call this method \textbf{\textit{Initial Learn}}, as the model learns on a small initial set of input feature vectors while predicting the rest.

In the following scenario, we retrain the model as specific intervals without considering concept drift. We can liken this to retraining a fraud prediction model every month or week so that the model is trained on the latest input data distributions. We call this method the \textbf{\textit{Regular Retrain}} method, as the model is regularly updated without considering concept drift. From a computational point of view, this is the most expensive model as the model is retrained far more often than required, irrespective of the fact whether a concept drift is present in the input data or not.

The final baseline scenario is where the model is continuously updated with the input data distribution. We can see this as partially updating a model as soon as the labels to the data are obtained. Although such a scenario is very unlikely in a real-world scenario due to the issues with updating models in a production environment and attaining labels for each and every data point. As explained by~\cite{hu2020no}, labelling data streams is expensive and requires quite a lot of effort to obtain. But such a baseline gives us an idea of how good a model could get with only partial updates. We call this baseline method the \textbf{\textit{Regular Update}}.

We compare all the methods using the Hoeffding Tree classifier with all default parameters. The \textbf{\textit{Regular Retrain}} method makes use of the reset method of the Hoeffding Tree classifier while the \textbf{\textit{Regular Update}} method uses the partial fit method of the algorithm.

\subsubsection{Results and Discussion:~~}

We set the training window size($\rho$) to one hundred input training instances based on experimentation with different values. Also~\cite{gozuaccik2019unsupervised} found that having a window size of 100 provides a good balance between generalizing to the current distribution and mostly having only a single distribution within the window. A smaller window size does not adequately generalise features for the network to discriminate between the different contexts adequately. 

Our method beats the state of the art in most datasets as seen in~\autoref{tab:results}. For larger datasets, like Rialto, Airlines, Poker or Covertype(marked with * in the table), we only considered the first 50000 input instances. This is due to the sheer number of unique drifts that arise and the increased training time once a previously unseen drift is detected. 

One of the best examples is for the \textbf{Airlines} dataset where our model did not detect a drift at all for the first 50000 input data instances. This also produces a better score compared to the other methods. But on the other hand, our model performs worse for the Electricity dataset. One reason for this could be that the dataset has a higher number of outliers within our window size($\rho$). 

\begin{table*}
\centering
    
   \caption{comparison of accuracy values of different methods.}
    \begin{tabular}{*6{c}}
    \toprule
    Dataset & Initial Train & Regular Update & Regular Retrain & OCDD & Our Method\\
    \midrule
    Electricity & 56.44 & 77.69 & 75.4 & \textbf{86.22} & 79.86\\
    Poker & 50.12 & 74.08 & 63.05 & 76.63 & \textbf{76.99*}\\
    Cover Type & 47.69 & 82.49 & 55.48 & \textbf{88.36} & 82.49*\\
    Rialto & 09.95 & 31.373 & 51.02 & \textbf{66.68} & 54.84\\
    Spam & 66.48 & 88.35 & 75.81 &  87.02 & \textbf{89.28} \\
    Phishing & 83.49 & 90.26 & 89.18 & 90.56 & \textbf{91.37}\\
    Airlines & 56.13 & 63.88 & 59.72 & 63.16 & \textbf{65.59*}\\
    Outdoor & 15.21 & 57.15 & 12.00 & 62.24 & \textbf{62.57*}\\ 
  \bottomrule
\end{tabular}
\label{tab:results}
\end{table*}

\section{Conclusion}\label{sec:conclusion}
Our method addresses the issue of identifying concept drifts that have occurred in the past, and we use that information to increase the training data for the secondary classifier. This memory of previous distributions allows our method to improve prediction accuracy, reduce the training time for the model, and reduce the effort in labelling data. Our model shows better performance when compared to other state-of-the-art models across the different publicly available datasets. Our method as a framework is easy to use and can be applied to different application domains with little to no modifications. We applied our method in an active learning loop to sample the most informative orbits to train a classifier. Our method performed vastly superior over random sampling and an active learning method. The only drawback of our framework is that the amount of training time increases with the increase in the number of unique drifts. Having to store the data of previous distributions in memory to retrain the model could also be memory inefficient. As for future work, we expect to use a Conditional Generative Adversarial Network that will enable us to simulate different input probability distributions effectively.

\bibliographystyle{bibliography/splncs04}
\bibliography{bibliography/bibliography}
\end{document}